\documentclass{article}

\usepackage{arxiv}

\usepackage[utf8]{inputenc} % allow utf-8 input
\usepackage[T1]{fontenc}    % use 8-bit T1 fonts
\usepackage{hyperref}       % hyperlinks
\usepackage{url}            % simple URL typesetting
\usepackage{booktabs}       % professional-quality tables
\usepackage{amsfonts}       % blackboard math symbols
\usepackage{nicefrac}       % compact symbols for 1/2, etc.
\usepackage{microtype}      % microtypography
\usepackage{lipsum}
\usepackage{graphicx}
\usepackage{subfig}

\title{Efficient Hyperparameter Optimization in Deep Learning Using a Variable Length Genetic Algorithm}

\author{Xueli Xiao\\
  Computer Science Department\\
  Georgia State University\\
  Atlanta, GA 30303 \\
  \texttt{xxiao2@student.gsu.edu} \\
  \And
  Ming Yan \\
  Institute of High Performance Computing\\
  National University of Singapore\\
  Singapore \\
  \texttt{yanmingtop@yahoo.com} \\
  \And
  Sunitha Basodi \\
  Computer Science Department\\
  Georgia State University\\
  Atlanta, GA 30303 \\
  \texttt{sbasodi1@student.gsu.edu} \\
  \And
  Chunyan Ji \\
  Computer Science Department\\
  Georgia State University\\
  Atlanta, GA 30303 \\
  \texttt{cji2@student.gsu.edu} \\
  \And
  Yi Pan \\
  Computer Science Department\\
  Georgia State University\\
  Atlanta, GA 30303 \\
  \texttt{yipan@gsu.edu} \\
}

\begin{document}
\maketitle

\begin{abstract}
Convolutional Neural Networks (CNN) have gained great success in many artificial intelligence tasks. However, finding a good set of hyperparameters for a CNN remains a challenging task. It usually takes an expert with deep knowledge, and trials and errors. Genetic algorithms have been used in hyperparameter optimizations. However, traditional genetic algorithms with fixed-length chromosomes may not be a good fit for optimizing deep learning hyperparameters, because deep learning models have variable number of hyperparameters depending on the model depth. As the depth increases, the number of hyperparameters grows exponentially, and searching becomes exponentially harder. It is important to have an efficient algorithm that can find a good model in reasonable time. In this article, we propose to use a variable length genetic algorithm (GA) to systematically and automatically tune the hyperparameters of a CNN to improve its performance. Experimental results show that our algorithm can find good CNN hyperparameters efficiently. It is clear from our experiments that if more time is spent on optimizing the hyperparameters, better results could be achieved. Theoretically, if we had unlimited time and CPU power, we could find the optimized hyperparameters and achieve the best results in the future.
\end{abstract}

% keywords can be removed
\keywords{Genetic Algorithms \and Hyperparameter Optimization \and Deep Learning \and Convolutional Neural Networks}

\section{Introduction}
Convolutional Neural Networks (CNNs) has achieved great performance in the area of computer vision. Designing the architecture of CNNs so that it produces good results on a task is challenging. It involves decisions on various hyperparameters of the network, such as how many layers are in the network, how many filters are in a layer, what are the sizes of convolutional windows, and so on. All the different combinations of the hyperparameters create a huge number of possibilities for CNN models. Traditionally, the selection of hyperparameters is done by hand tuning, ie. manually testing different sets of parameters that may work well. This requires certain level of knowledge and expertise in deep learning, and the process of trial and error is tedious, time-consuming, and may be discouraging to the novices. It is thus beneficial to automatically search for a CNN architecture that suits the needs of the task. Searching for a good deep learning model is challenging, very computationally intensive, and time consuming. The lengthy search time is caused by many factors: the evaluation time for a single deep learning model is long, the search space grows exponentially large as the model depth increases, and the access to GPU resources is limited -- many researchers only have access to one or a few GPUs. Given the above mentioned reasons, it is certainly not practical to traverse every single possibility to find the best model.

Evolutionary algorithms have been used in hyperparameter optimization of deep learning models. Young et al. \cite{Young2015OptimizingAlgorithm} used a genetic algorithm (GA) to optimize the hyperparameter of a 3-layer CNN. The algorithm is not suitable for situations where we do not know the how many layers are needed there. Real et al. \cite{Real2017Large-ScaleClassifiers} used a mutation only evolutionary algorithm, and gradually grows the deep learning model to find a good set of combinations. The evolutionary process is slow due to the mutation only nature.

Inspired by the work of \cite{Young2015OptimizingAlgorithm} and \cite{Real2017Large-ScaleClassifiers}, we propose a variable length genetic algorithm to efficiently optimize the hyperparameters in CNNs. Models of various hyperparameters settings are first created and evolved using the algorithm. We do not put a constraint on the model depth, and various techniques such as crossover operation in the evolving process makes the algorithm more efficient. Experimental results show that our method can efficiently find good hyperparameter combinations even when the search space becomes exponentially large. Hyperparameter optimization is itself a time consuming process. Having an efficient algorithm is especially helpful when the access to GPU hardware is limited and the search space is very large. The major contribution of this work is as follows:
\begin{itemize}
    \item Our method can efficiently discover deep learning models with variable depths. It makes hyperparameter optimization more feasible for researchers with limited computing resources.
    \item Our algorithm does not have a constraint on the depth of deep learning models, which means our method can handle problems of different sizes.
    \item We incorporated crossover operation in our model evolution process along with efficient model evaluation techniques to make the model discovering process more effective. 
\end{itemize}
The rest of the work is organized as follows: Section II describes related work in hyperparameter optimization. Section III introduces our method in searching for the best hyperparameter combinations. Section IV includes experiment details and settings. Section V is conclusion and possible future research directions.

\section{Background}
\subsection{Convolutional Neural Networks}

A Convolutional Neural Network (CNN) is a type of deep artificial neural network and it is assumed to be used for images. Classical CNNs contain convolutional layers, pooling layers, ReLU activations, and fully connected layers. Convolutional layers have convolutional windows that can extract features from images. These windows are usually called filters or kernels. The kernels move through the images, performing convolutions on local image data and produce feature maps. 

Pooling is used to reduce the number of parameters in a CNN, it is also called downsampling or subsampling. It can help reduce computation complexity. If a feature map is of size 2020, using pooling of size 22 and step of size 2, the dimension can be reduced to 1010 .There are different types of pooling layers: max pooling, average pooling, fractional max pooling and so on. Max pooling takes the pixel of the biggest value in the pooling window, whereas average pooling calculates the average of all the pixels in the window. These are the two commonly used pooling methods in convolutional neural networks. Activation function ReLU stands for Rectified Linear Unit, it introduces nonlinearity into the network. ReLU activation keeps all positive input and makes negative input to be zero: 
f(x)=Max(0,x).
Some of the other types of activation functions include tanh, Exponential Linear Unit (ELU) \cite{Clevert2015FastELUs}, Scaled Exponential Linear Unit (SELU) \cite{Klambauer2017Self-NormalizingNetworks} and so on. 
Fully connected layer has full connections to neurons in the previous layer. 

\subsection{Genetic Algorithms}
Genetic algorithms are biologically inspired. It learns from Darwin's evolutionary theory. In Darwin's theory of natural selection, the fittest individuals are selected and they produce offsprings. The characteristics of these individuals are passed on the next generations. If the parents have higher fitness, their offsprings tend to be fitter and have more chances of survival. 

Genetic algorithms learn from this idea. They can be used to solve optimization and search problems. In genetic algorithms candidate solutions are evolved to generate better ones.  A set of solutions form a search space and the goal is to find the best ones among them. This is similar to finding the fittest individual in a population. Genetic algorithms start with a population that contains random solutions within the search space. 

Each solution has a chromosome, which contains properties about the solution. These chromosomes can be altered. A typical genetic algorithm has three bio-inspired operators that can be used on a chromosome: selection, crossover, and mutation. Selection means to select a portion of the population as candidates to produce offsprings and generate more solutions. Usually the fitter individuals are selected. The fitness of a solution can be calculated using a fitness function, and it reflects how good the solution is. Crossover combines the chromosomes of two parents, and produces a chromosome for the offspring. The offspring's chromosome inherits the properties from both parents. The Mutation operator is like biological mutation: it changes one or more values in the chromosome. This introduces more diversities in the population. 

Genetic algorithms are simple yet effective. They have been applied on various research problems such as vehicle routing problems \cite{Baker2003AProblem}, dynamic channel-assignment problems \cite{Fu2006UsingProblem}, neuroscience search problems \cite{Xiao2018AnModel} \cite{Xiao2017DetectingModel}, deep learning hyperparameter optimizations \cite{Young2015OptimizingAlgorithm} \cite{Real2017Large-ScaleClassifiers}, neural network weight optimizations \cite{Gupta1999ComparingTraining} \cite{Montana1989TrainingAlgorithms} and so on. Genetic algorithms with variable chromosome length have also been applied to many problems \cite{Pawar2015GeneticDetection} \cite{Srikanth1995AClassification}. 

\subsection{Hyperparameter Optimization Algorithms}
Grid search and random search \cite{Bergstra2012RandomOptimization} are two popular methods for hyperparameter optimization. Grid search selects the best model among many models built on predefined hyperparameter settings. It evaluates all the models to guarantee the best one is found. As the number of hyperparameters grow, and search space gets increasingly larger, the amount of time it takes for grid search grows exponentially, making it impractical to be used. Random search \cite{Bergstra2012RandomOptimization}takes less time than grid search, because unlike grid search, it does not search all possibilities exhaustively. Instead of trying out all possibilities, it randomly selects models with different hyperparameter combinations. There is a trade-off between search time and quality of discovered models. It is not guaranteed that the optimal hyperparameter combination within the search space will be found. It turns out random search can achieve similar results compared with grid search, while being more efficient. 

Besides grid search and random search there are other approaches, such as Bayesian optimization, gradient based methods, reinforcement learning based methods, and evolutionary algorithms. Bayesian optimization methods \cite{Snoek2012PracticalAlgorithms} \cite{Shahriari2016TakingOptimization} \cite{Swersky2013Multi-TaskOptimization} use an acquisition function to smartly decide where should the algorithm explore next in the search space. A probabilistic model is built to balance exploration and exploitation. Gradient based methods can be used if the hyperparameters are continuous. Luketina et al.\cite{Luketina2016ScalableHyperparameters} and Fu et al. \cite{Fu2016DrMAD:Networks} used gradient descent to optimize hyperparameters. Reinforcement learning based methods are used to automatically discover CNN architectures. Zoph et al. \cite{Zoph2017NeuralLearning} used a Recurrent Neural Network to generate CNN model structure, and applied reinforcement learning to improve the generated architecture. Baker et al. \cite{Baker2016DesigningLearning} used a Q-learning agent to discover a network layer by layer. 

Another large category of hyperparameter optimization method is based on evolutionary algorithms. Research work under this category differs in their evolutionary algorithms used, the search space, the encoding scheme for CNN model architecture, ways of reducing computational cost and so on.  Young et al. \cite{Young2015OptimizingAlgorithm} performed genetic algorithms to optimize hyperparameters in CNNs. A three layer fixed-architecture CNN is used, and only six hyper parameters are tuned. The problem with a fixed layer architecture is that the models may be too small to fit the problem, leading to a high bias. Real et al.\cite{Real2017Large-ScaleClassifiers} evolved complex CNN architectures from very simple individuals by applying different kinds of mutation operations. Model structures are encoded as graphs in \cite{Real2017Large-ScaleClassifiers}. While the method by Real et al.\cite{Real2017Large-ScaleClassifiers} can find accurate models for challenging tasks, the computational cost is very high: 250 workers are used and the searching time is over 250 hours. A mutation only algorithm can contribute to high computational cost since the evolutionary process is very slow. Many hyperparameter optimization methods use the model’s accuracy on the held out validation dataset as an evaluation of the model. Albelwi et al.\cite{Albelwi2017ANetworks} took a different approach of evaluating how good a CNN model is. Images are reconstructed from the learned filters using deconvnet, and the similarity between the original image and the reconstructed image is used as the fitness of a CNN architecture. There are various ways of defining models’ structures too. Suaganuma et al.\cite{Suganuma2017AArchitectures} used Cartesian genetic programming encoding scheme to define CNN models. It enables them to incorporate modules such as tensor concatenation into the model architecture. Baldominos et al.\cite{Baldominos2018EvolutionaryRecognition} used backus-naur form to define a grammar for model topologies, allowing more flexible representations of CNN architectures.

\section{Method} \label{sec:method}

Evolutionary algorithms (EA) are inspired by the process of natural selection. They are commonly used to produce high-quality solutions when the search space is large. The idea of EA is to produce an initial population that is made up of diverse solutions, and evolve them for many generations to produce better ones. The genes of the better solutions will survive and pass on to further generations, whereas the worse solutions die off.  

Both Young et al. and Real et. al use EA, and there are limitations. For the purpose of optimizing hyperparameters for deep learning models, no constraints should be put on the network layers. Limiting the number of layers could make the network size too small for the problem, leading to underfitting. Young et. al used a classical fixed length genetic algorithm (GA), the most popular type of GA, to optimize a CNN model for the CIFAR-10 dataset \cite{Krizhevsky2009LearningImages}. A fixed length GA means the CNN model has predefined fixed number of layers. However, for a given problem, one never knows how many layers are enough before experiments are done. Having a too large network on the other hand, can lead to overfitting. Thus it makes sense to start from a small network, and gradually grow it as needed. Real et. al do so by using a mutation only EA. Although high accuracies are obtained on the CIFAR-10, and CIFAR-100 dataset, the computational cost of their method is very high. The whole process takes 250 parallel computers and over 250 hours to complete. Those who rely on automatic hyperparameter optimization tools usually do not have that kind of computational resource. 

Because of the above mentioned limitations, we propose a variable length genetic algorithm for deep learning hyperparameter optimization. It starts from a small network model and gradually builds on top of it. Crossover operation along with other techniques are used in this EA to accelerate the optimization process, making it affordable for those with limited resources.

\subsection{Variable Length Genetic Algorithms}
Classical GA requires fixed chromosome length. Since the number of convolutional layers varies in different CNN models, and more hyperparameters are involved as the model grows deeper, a variable length GA is more suitable for the purpose of our task.

A chromosome in the variable length GA contains parameters that define a solution, in this case, a hyperparameter configuration for a network model. How the solution is encoded in the chromosome is discussed in section 3.2. 
  
\begin{figure}
  \centering
  \includegraphics[width=1.0\columnwidth]{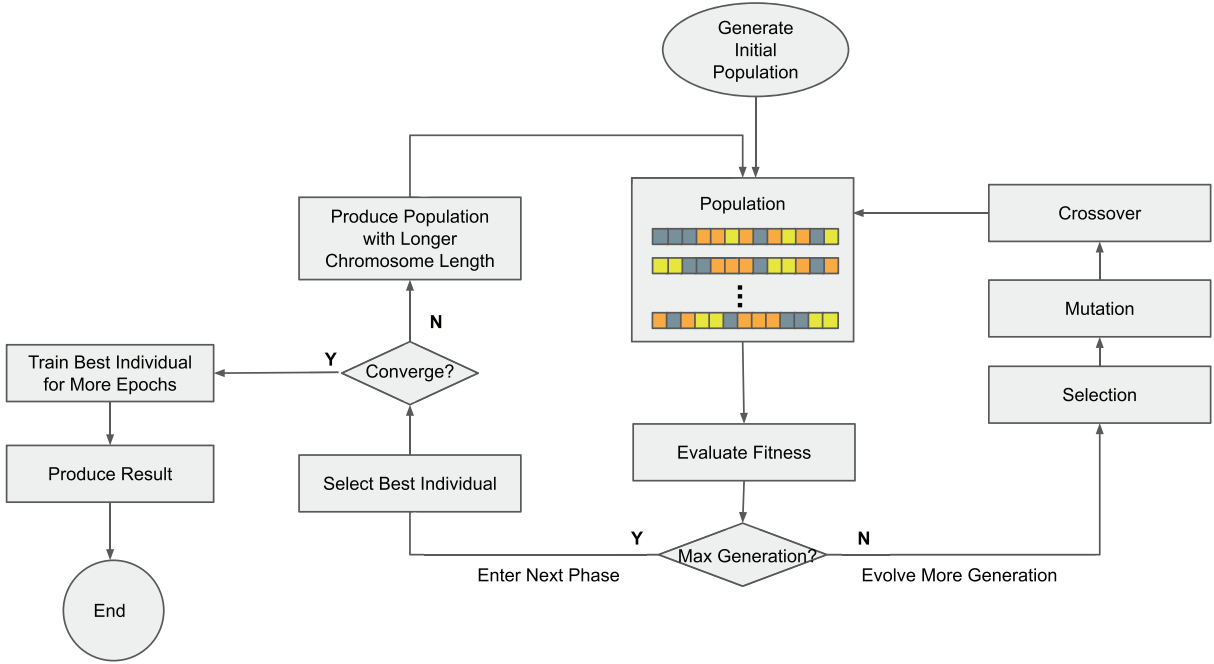}
  \caption{The hyperparameter optimization steps using our variable length genetic algorithm.}
  \label{GA}
\end{figure}

Figure \ref{GA} describes the overall procedure for the variable length GA. The initial population with two convolutional layers is first produced, and the individuals/solutions are initialized with random hyperparameters. The individuals are then evaluated and sorted based on how fit they are. The fitness of an individual is made to be the corresponding CNN model’s accuracy on the validation dataset. Then the fitter individuals gets directly picked, and survive into the next generation. The rest of the next generation are produced by individuals from the current generation using crossovers and mutations, common operations of a GA. In crossover, two individuals are selected as parents to produce a child that inherits the parents’ properties. In mutation, some properties of the individual are changed to other values. After the 2-layer models have been evolved for some generations, the algorithm enters the next phase that enables more layers and hyperparameters. 

A new phase in the algorithm simply means the chromosome length increases, bringing more hyperparameters in, and the corresponding CNN models have more layers. 

The steps of our hyperparameter optimization method are as follows:
\begin{enumerate}
\item Initialize the number of populations $p$ and generations $g$ according to user input.
\item Create the initial population. Each individual in the population has randomized hyperparameters. The size of the population is $p$ as specified by the user. 
\item Evaluate the fitness of each individual in the current generation. The fitness is the accuracy of the individual on the validation dataset. 
\item Sort each individual according to their fitness value, from high to low.
\item Select the fittest individuals. Select a portion of the population that has the highest fitness values, and let them survive into the next generation. The percentage of this survival rate can be set manually. 
\item Allow some less fit individuals to survive. For the rest of the individuals in the current population, give them a small chance to survive into the next generation. This probability is set manually.  
\item Randomly mutate some individuals in the next generation. The mutation rate is manually configured. If an individual is chosen to mutate, one of its hyperparameter’s value will be changed. 
\item Produce new individuals. Randomly select two individuals from the next population to serve as parents, and perform crossover operation to produce a child. Each hyperparameter in the child is randomly set to be one of its parent’s hyperparameter value. Repeat the crossover operation on different random individuals many times until the next generation has $p$ individuals. Now the next generation becomes the current generation.
\item Repeat Step 3 - 8 until the number of generations has reached g.
\item Select the best individual from the current population.
\item Produce population with longer chromosome length. For each individual of the population, one part of the chromosome is from the best individual in Step 10, the other part is randomly generated. 
\item Repeat step 9 - 11 until convergence. 
\item Select the best individual and train for more epochs until convergence.
\end{enumerate}

% \begin{figure} % picture
%     \centering
%     \includegraphics{test.png}
% \end{figure}

\subsection{Encoding Scheme}
The hyperparameter information of each individual, or CNN model, is encoded into what’s called a chromosome in GA. The chromosome may have different representations, such as a bit string that contains 1’s and 0’s. For the purpose of hyperparameter optimization, the chromosome of our experiments have many fields that contain different values. It defines the hyperparameters of a CNN architecture, such as the number of neurons, activation function type, and so on. Once a chromosome is known, a CNN model can be built according to the chromosome. 
 
\begin{figure}[t]
\centerline{\includegraphics[width=1.0\columnwidth]{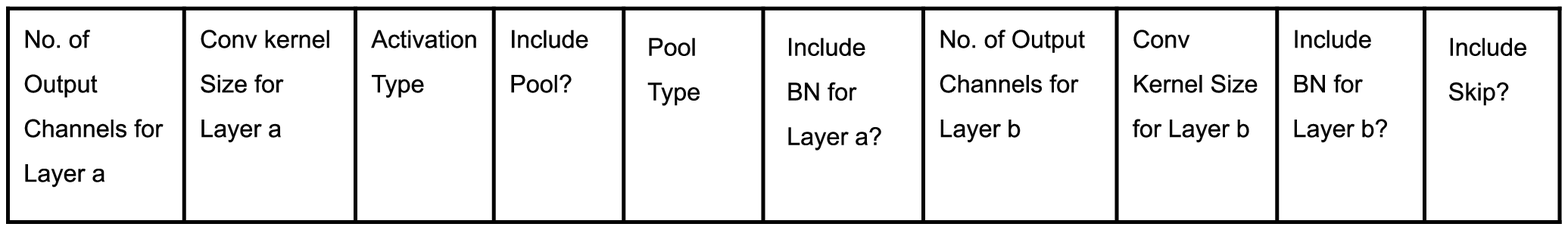}}
\caption{The initial chromosome (Phase 0). In the chromosome, what type of activation function to be used for the whole model is encoded. It also includes hyperparameter information for the first two convolutional layers: the size of the convolutional windows, number of output channels, and whether or not to include batch normalization for the layer. For the whole two-layer block, there are three additional hyperparameters: whether to add a pooling layer after the first convolutional layer, what is the pooling type, and if a skip connection should be included.}
\label{phase0_encoding}
\end{figure}

\begin{figure}[t]
\centerline{\includegraphics[width=0.3\columnwidth]{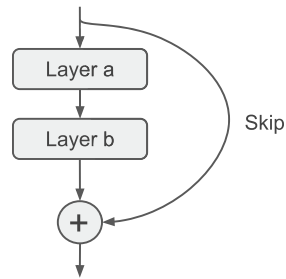}}
\caption{An example of a skip connection. This block contains two convolutional layers: Layer a and Layer b. The skip connection performs a 1x1 convolution on the input, and its output is added with the output of Layer b. Stride size and number of filters of the skip connection is setup such that its output tensor has the same shape as that of Layer b.}
\label{skip}
\end{figure}

In the initial phase (Phase 0) of the algorithm, there are two convolutional layers a and b. Relevant hyperparameters are encoded in the chromosome as shown is Figure \ref{phase0_encoding}. For each convolutional layer, the hyperparameters include the number of output feature maps, the size of the convolutional windows, and whether or not to include batch normalization. For this two-layer block, there are three additional hyperparameters: whether to include pooling in this block, what is the pooling type, and whether or not to include a skip connection. If there is a pooling layer for this block, it is going to be added after the first convolutional layer. If a skip connection exists, it is going to have a 1x1 convolution, and its output is going to be added with the output of the whole block. Figure \ref{skip} is an example of the skip connection.  There is an additional “activation type” field in the chromosome for Phase 0. This defines what type of activation function is going to be used for the whole model.  Figure \ref{phase0_encoding} is the encoding scheme of Phase 0, the initial phase that represents a model with two convolutional layers. 

\begin{figure}[t]
\centerline{\includegraphics[width=1.0\columnwidth]{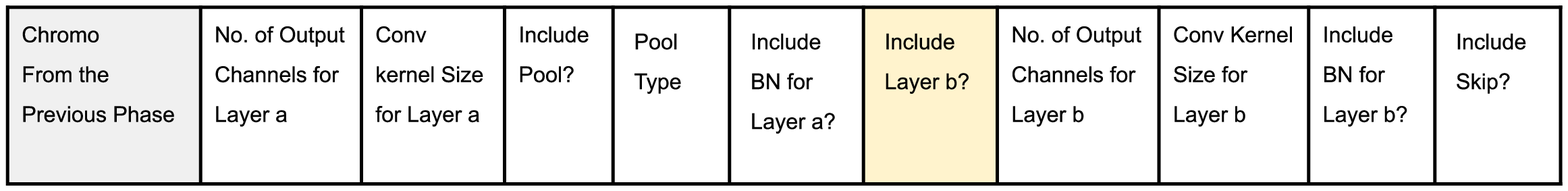}}
\caption{Encoding for Phases after Phase 0. Chromosome in the current phase is an extension of the one from the previous phase plus some new fields. Chromosomes are longer, and the represented models have more layers. The field with gray background indicate the chromosome from the previous phase. The rest defines the properties of new layers in the current phase. In the current, two convolutional layers can be added, with the flexibility of letting the algorithm to decide whether to add one or two. This is achieved by adding a “Include Layer b?” field, the one with yellow background, in the chromosome.}
\label{phasen_encoding}
\end{figure}

For every phase after Phase 0, we allow the algorithm to still add two convolutional layers (Layer a and Layer b) in each phase, but we give the flexibility and let the algorithm decide whether to add one or two convolutional layers in the phase. The chromosome encoding for a Phase after Phase 0 is as shown in Figure \ref{phasen_encoding}. New fields are concatenated to the chromosome of the previous phase. The new part is very similar to the encoding for Phase 0, except that it does not have an “Activation Type” field, and has a new field “Include Layer?” that can take two values: 0 or 1. If the value is 1, there will be two convolutional layers in this phase, and if 0, there will be only one. 

\begin{figure}[t]
\centerline{\includegraphics[width=0.7\columnwidth]{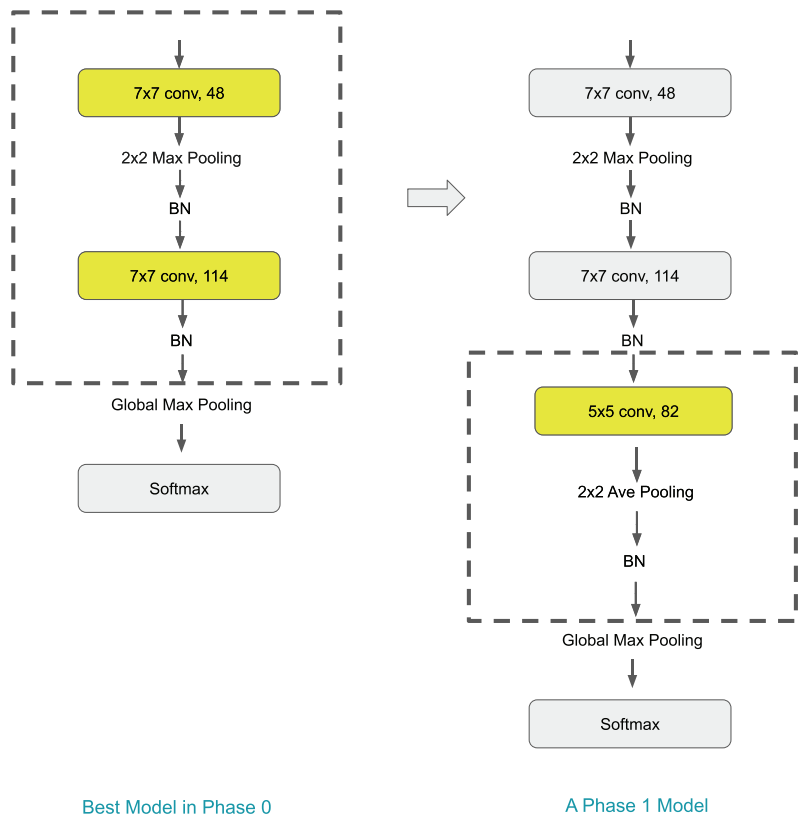}}
\caption{The model grows deeper when a new phase begins. The models in the new phase are built based on the best model in the previous phase. To the left is an example of the best model found in Phase 0, and to the right is a model in Phase 1.}
\label{new_phase}
\end{figure}

\begin{figure}[t]
\centerline{\includegraphics[width=1.0\columnwidth]{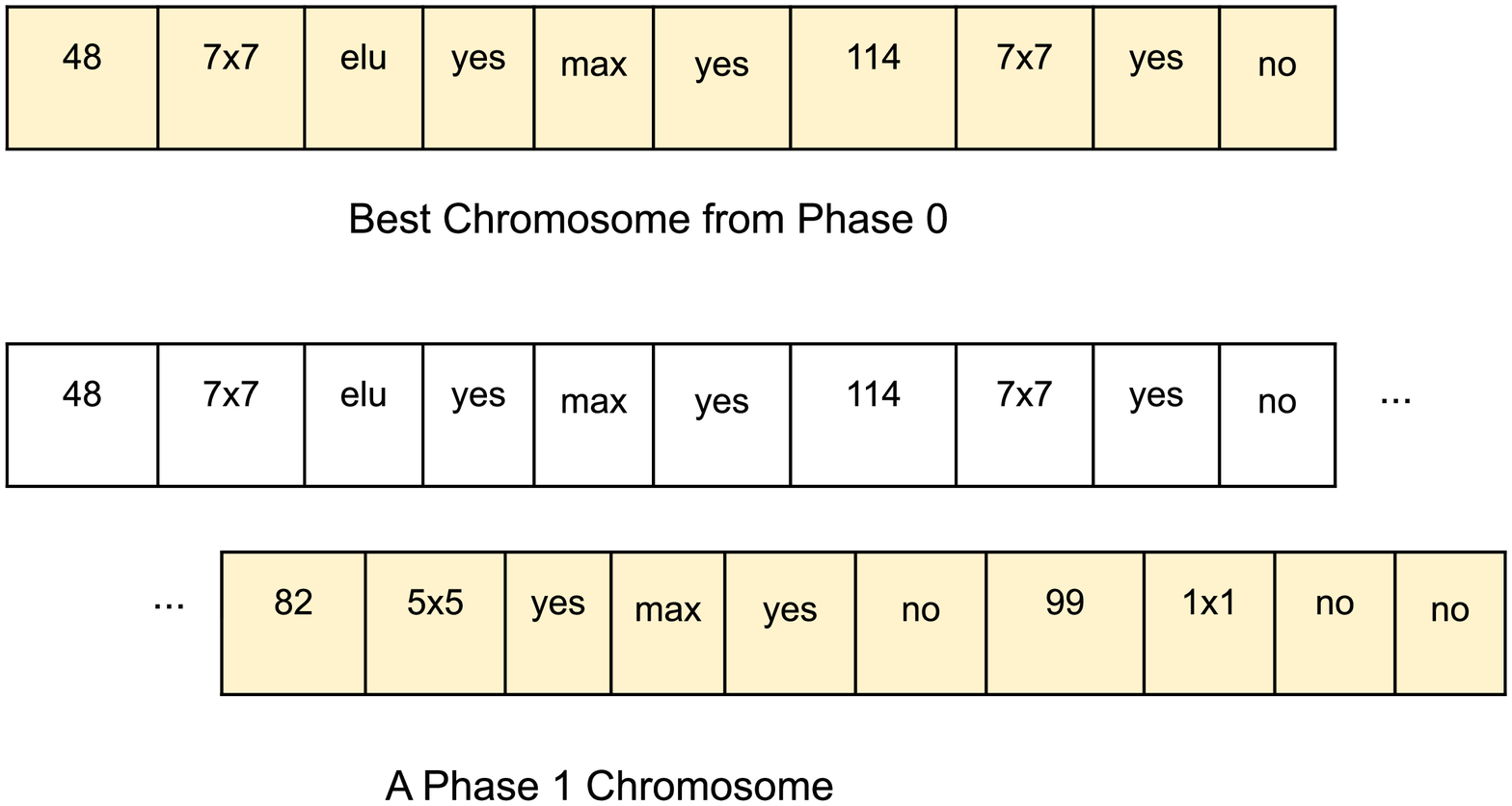}}
\caption{The corresponding chromosomes for models in Figure \ref{new_phase}. Chromosomes grows longer when a new phase is entered. The chromosome in the new phase is extended from the best chromosome in the previous phase.}
\label{new_phase_chromo}
\end{figure}

Going into a new phase means that the chromosomes are longer, and models being discovered by the algorithm grow deeper. And part of the hyperparameters in the deeper model comes from the best model in the previous phase. An example can be found in Figure \ref{new_phase}. Assume to the left is the best model discovered in Phase 0. The part in the dashboard box is encoded in Phase 0 chromosome. When the algorithm enters Phase 1, the chromosomes grows longer, and the corresponding CNN models are deeper. The deeper models share some common properties: part of their hyperparameter settings come from the best model in Phase 0. See the right part in Figure \ref{new_phase} for an example of Phase 1 model. The part bounded by dashed box is being searched in Phase 1, and the gray part above is from Phase 0. Figure \ref{new_phase_chromo} shows the corresponding chromosomes for the two models in Figure \ref{new_phase}. In Figure \ref{new_phase_chromo}, the yellow fields indicate hyperparameters that are being searched in that phase, and gray fields indicate hyperparameters from previous phases. 

\subsection{Fitness of individuals}
The fitness of an individual is its accuracy on the validation dataset. One way is to train the models to converge, and then compare. While this is more accurate in terms of comparing the models’ performance on the test set, this process takes very long time and needs a great amount of computing power. To save time, models can be trained for just a few epochs to compare their relative fitness with each other. Models that are fitter in the first few training epochs tend to also perform better later. In the experiments, the number of training epochs for comparing relative fitness is set to be 5. 

\begin{figure}[t]
\centerline{\includegraphics[width=0.8\columnwidth]{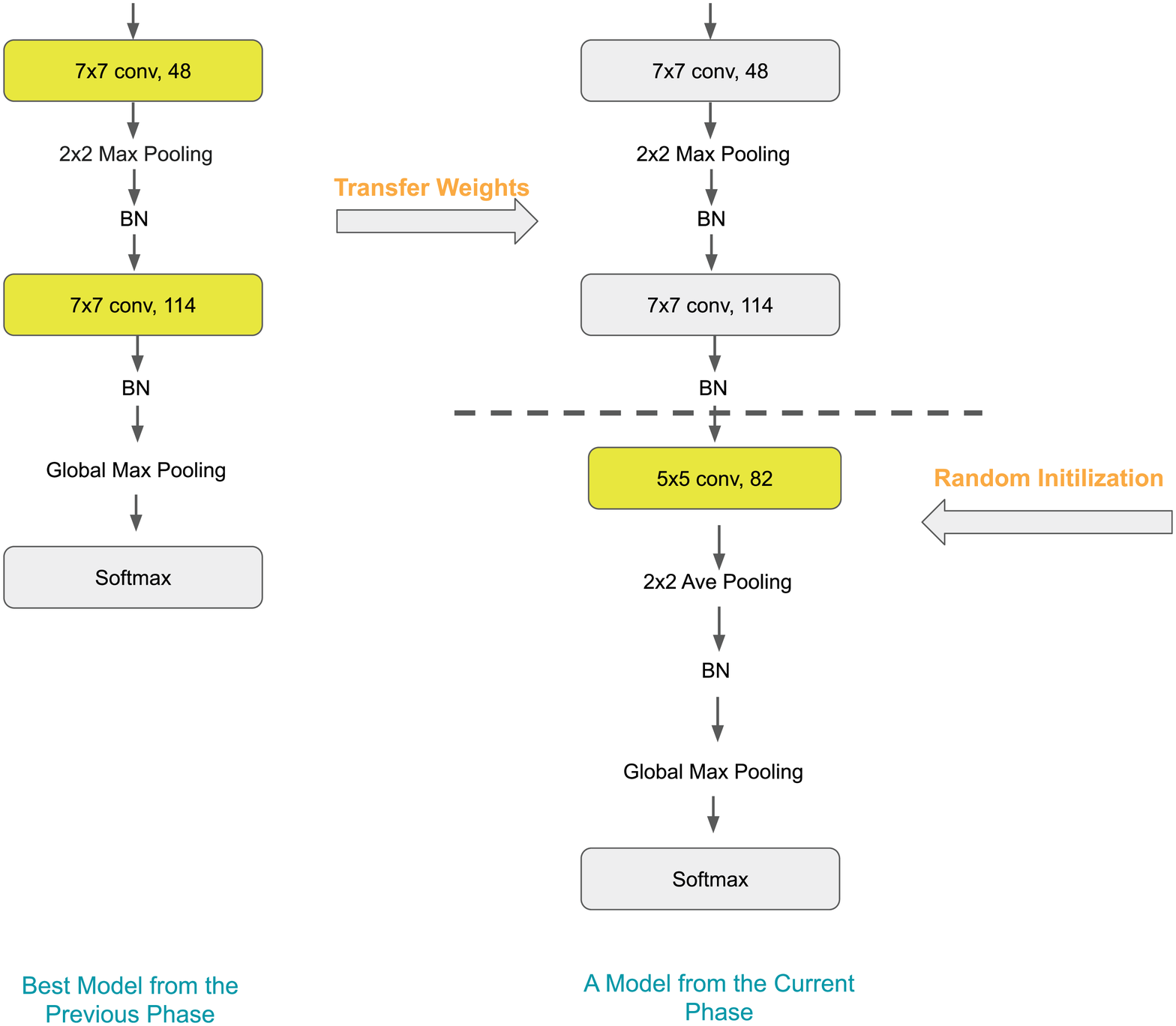}}
\caption{Weight initialization scheme for deeper models. A deeper model of the current phase is built on top of the best model from the previous phase. Part of its weights is transferred from the previous best model, part is initialized randomly.}
\label{transfer_weights}
\end{figure}

This way of evaluation is unfair when the size of the model differs greatly. A deeper model might have similar or even worse test accuracy compared with a shallower model when both are only trained for a small number of epochs such as five. To solve this problem, we introduce weight transfer from shallow models to deeper ones. In Figure \ref{transfer_weights}, when the algorithm enters a new phase, it searches for deeper models (the right part). New layers are added to the best model from the previous phase (the left part). Instead of initializing all the weights randomly in the new models, some of their weights are transferred from the previous best model, and some are done through random initialization. This takes advantage of the previous trained models and saves some training time. To evaluate its fitness, the deeper model is simply trained for five epochs, since part of it already received some training.

\section{Experiments and Results} \label{sec:experiments}

\subsection{Datasets}

Experiments were done on the CIFAR-10 dataset \cite{Krizhevsky2009LearningImages}. This dataset contains 60000 color images of size 32X32 in 10 classes: each class has 6000 images. There are 50000 images for training, and 10000 for testing.

\subsection{Search Space}
The search space defines all the possible solutions that can be searched. A relatively small search space may take the algorithm less time to find a satisfying CNN model, but may be constrictive due to the limited number of possible models included in this space. A larger one, on the other hand, is more time consuming and requires more computational power. 

\begin{figure}[t]
\centerline{\includegraphics[width=0.6\columnwidth]{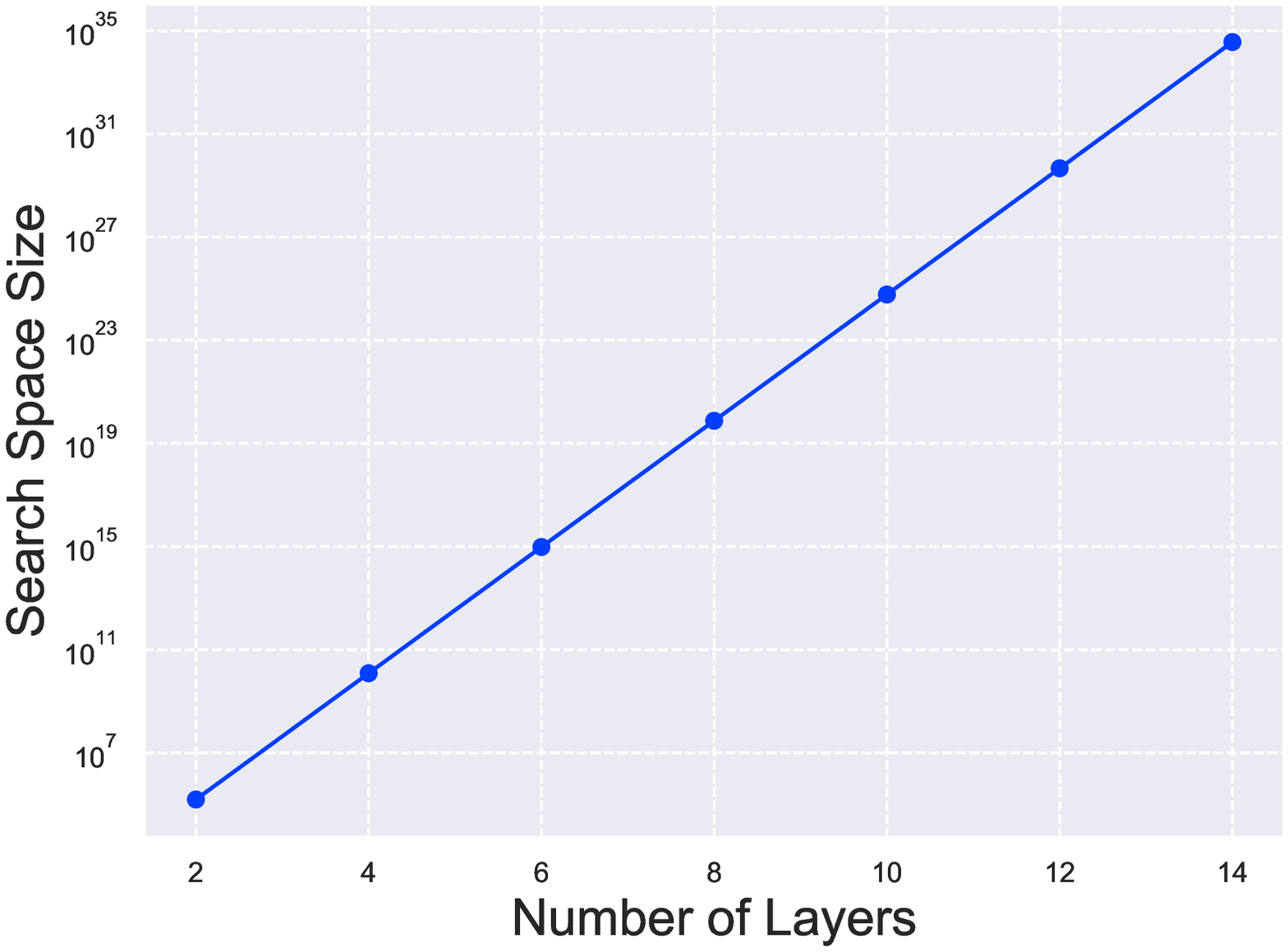}}
\caption{The size of the search space grows exponentially as the number  of layers in the deep learning model increases. In the initial phase of our algorithm, models have two layers and the search space is 156,800. As the model goes deeper, this number grows exponentially. When the number of layers reaches 14, the search space grows to about $10^{35}$.}
\label{space_size}
\end{figure}

In our experiments, the search space includes the number of output feature maps in each convolutional layer, convolutional filter size, activation function type, pooling type, whether to have a skip connection, whether to include batch normalization in a layer, and total number of layers. Most of the hyperparameters, such as number of output feature maps and convolutional filter sizes need to be decided in every convolutional layer. Thus as the model grows deeper, the search space becomes exponentially larger, and the searching more difficult. Most of the previous works do not include pooling types in their search space. In our work however, various pooling layers can exist in a single model. Table~\ref{tab1} describes the default search space of our experiments. The left column are the types of hyperparameters and the right column are their choices. Unless otherwise noted, our variable length genetic algorithm uses this space. Figure \ref{space_size} shows how the search space of algorithm grows as the number of layer increases. Initially models have two layers and the search space contains 156,800 different combinations of hyperparameters. When the the number of layers increases to 14, the search space grows to about $10^{35}$.

\begin{table}[htbp]
\begin{center}
\caption{The default search space used by our variable length genetic algorithm.}
\begin{tabular}{cc}
\toprule
  \textbf{Hyperparameter} & \textbf{Choices} \\
\hline
 Number of & 8, 16, 32, 64, 128, 256, 512\\
 Output Channels & \\
 \hline
Convolutional & 1x1, 3x3, 5x5, 7x7, 9x9 \\
Filter Size & \\
\hline
Activation Function & ReLU, Tanh, ELU, SELU\\
Type & \\
\hline

Pooling & Max pooling, Average pooling\\
 Type &\\
 \hline
 
 Skip & Yes, No\\
 Connection &\\
 \hline
 
 Batch & Yes, No\\
 Normalization &\\
 \hline
 
 Number of & $\geq 2$ \\
 Layers &\\
\bottomrule

\end{tabular}
\label{tab1}
\end{center}
\end{table}

The possible options of each hyperparameter type are carefully chosen. ReLU, Tanh, ELU and SELU are possible choices for the activation functions. Tanh is chosen in favor of Sigmoid because it is zero centered.

\subsection{Experiment Setup}
Our proposed method is implemented in Python, and CNN model are created using Keras. Experiments were run on a single GPU. To save time, the fitness of each individual is made to be the test accuracy after five training epochs. Weight transfer is also used to save the experiment running time. If a model survives into a new generation, it does not need to be trained again because all of its weights are saved. If the algorithm enters a new phase where the new population with longer chromosomes are built on top of previous stage’s best individuals, the old best model’s trained weights are transferred into the new models, and the new ones are trained for another five epochs to get their fitness value. 

\subsection{Genetic Algorithm Settings}
There are parameters that need to be set up in our variable length genetic algorithm: the size of the population, number of generations, number of phases, mutation rate, and survival chance of less fit individuals. In the experiments, 50\% of the fittest individual survive into the next generation, the survival rate of less fit individual is set to be 0.2, and the mutation rate is set to be 0.2. We do not set a predefined number of phases in our method. Instead, a stopping condition is set to stop the algorithm. Section \ref{stop} discusses the stopping condition in details.

\subsection{Stopping Condition}\label{stop}
\begin{figure}[t]
\centerline{\includegraphics[width=0.6\columnwidth]{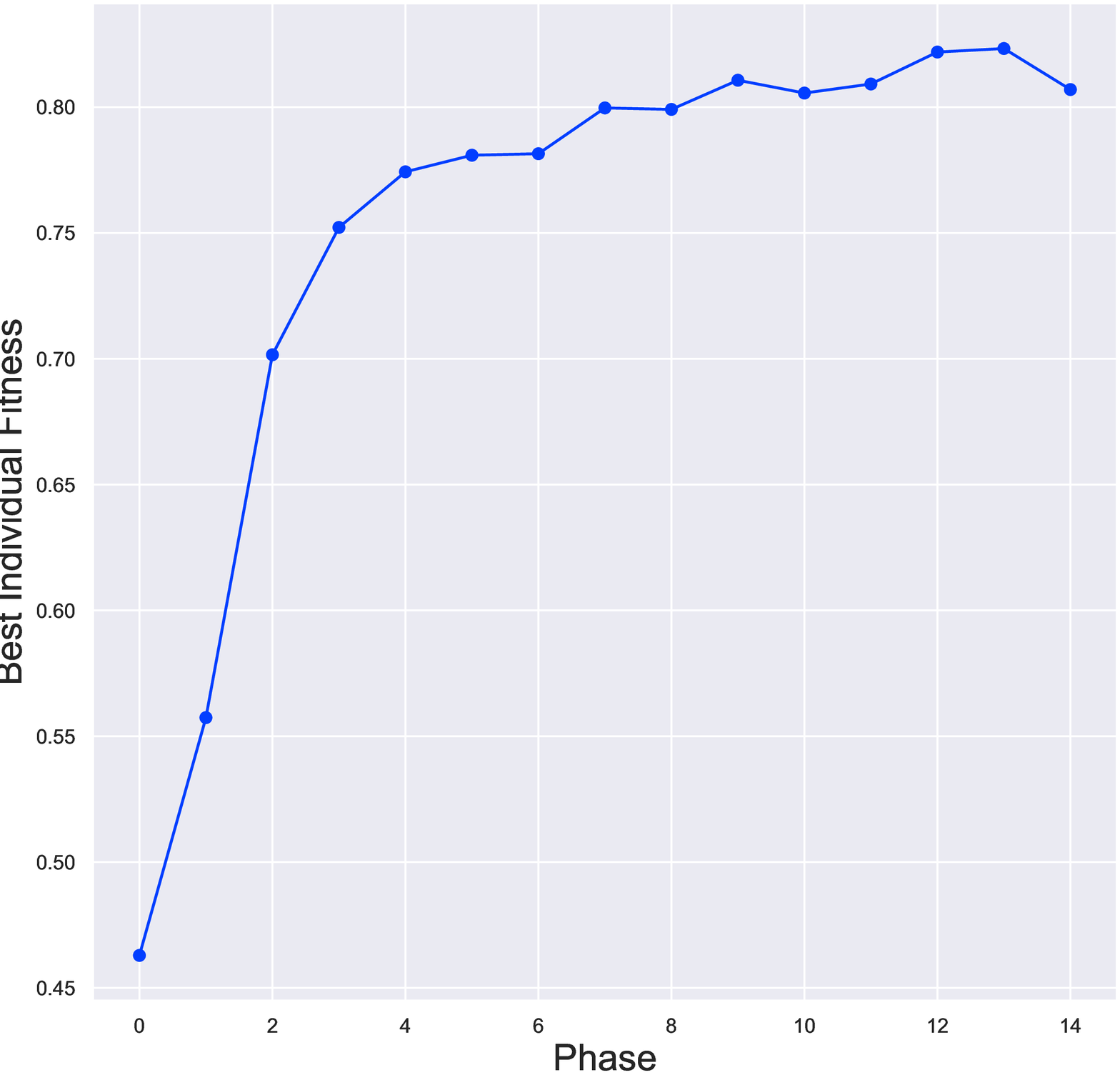}}
\caption{The evolutionary process stops when the stopping condition is met. In this example the algorithm stops at Phase 14 because the fitness drop is larger than the threshold. The best individual, in this case at Phase 13, is picked to receive further training.}
\label{stopping_condition}
\end{figure}

The evolutionary process stops when there is no more improvement on validation accuracy when more layers are added. A threshold value is set for the stopping condition: when the best fitness in the new phase is smaller than the best fitness in the previous phases by more than 0.01, the evolving process stops, and the individual with the best fitness is returned. This individual then receives further training. In Figure \ref{stopping_condition}, the evolutionary process stops at the 14th phase. The best individual fitness at Phase 14 is 0.8070, whereas the best in the previous phases is 0.8233 (at Phase 13). The evolutionary process thus stops, and the best individual at Phase 13 is picked to receive further training.

\subsection{Results}
\begin{figure}[t]
\centerline{\includegraphics[width=0.7\columnwidth]{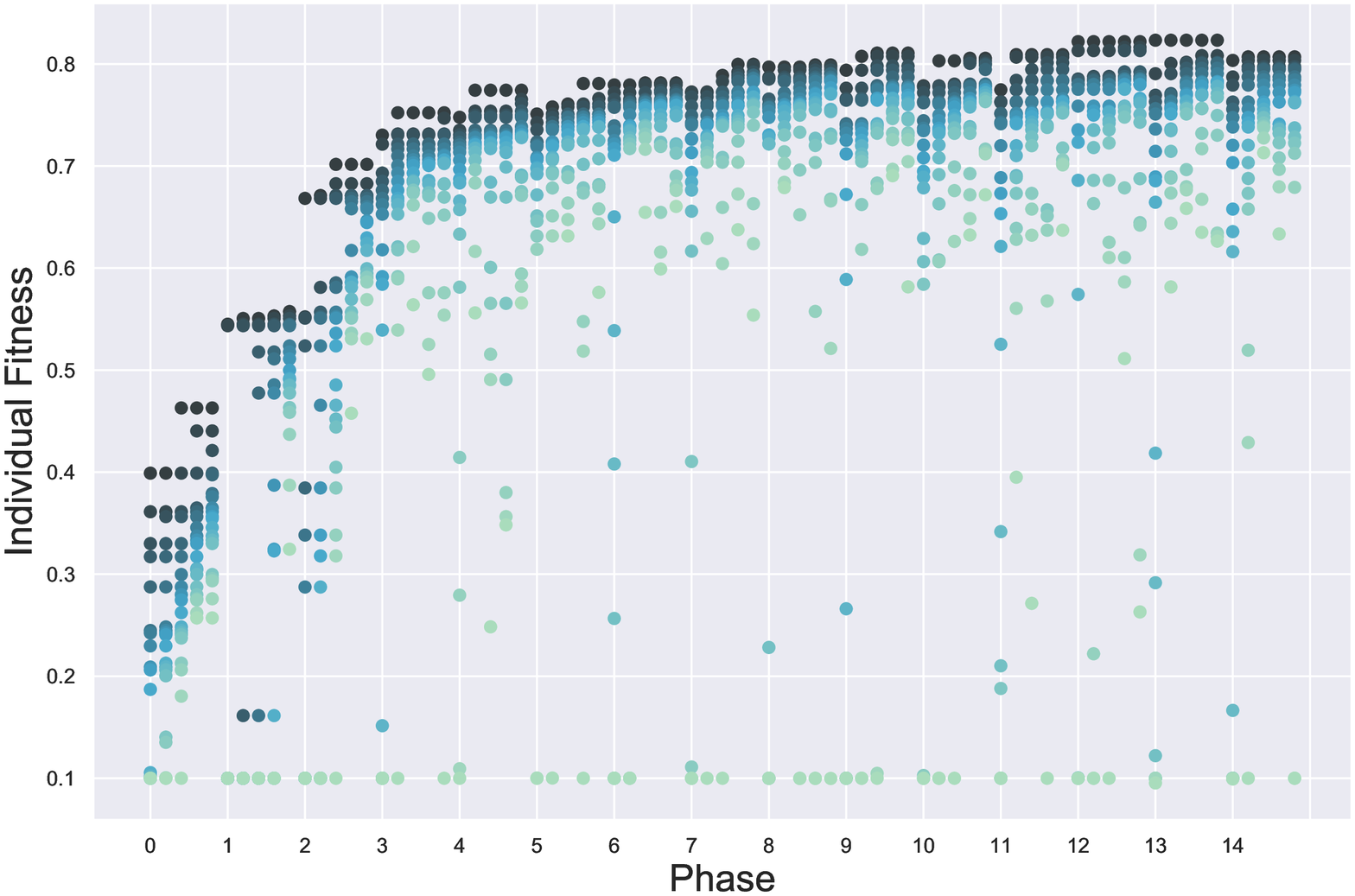}}
\caption{The fitness of 20 individuals evolved for 5 generations in each phase. The algorithm stops at Phase 14 when the stopping condition is met. The best individual found in this evolutionary process is chosen to receive more training. The overall evolution time is less than 25 hours.}
\label{evolve_1}
\end{figure}

Figure \ref{evolve_1} shows the result of 20 individuals evolved for 5 generations in each phase. Each individual is trained for 5 epochs to evaluate the fitness. The evolution stops at Phase 14, when the stopping condition is met. The number of convolutional layers increases in each phase. Initially, when the algorithm enters a new phase, there is a great improvement in the accuracy of the individuals, this improvement gradually declines in the later phases. This shows the increase of convolutional layers can have great influence on the model performance, especially when the model is relatively shallow. When the network is shallow, its size could be too small to fit the problem. As the number of layers increase, the size of the network grows, and so is the capability of the network. 

We compare the results of our variable length genetic algorithm with random search, classical genetic algorithm \cite{Young2015OptimizingAlgorithm}, and large scale evolution \cite{Real2017Large-ScaleClassifiers}.

Unlike our method that allows for a dynamic search space that grows as the number of layers increases, random search requires that the search space to be fixed, which means the maximum number of layers need to be specified. We make the number of layers for random search in the range $[2,10]$. Other search space settings are similar to our method.

\begin{figure}[t]
\centering
  \subfloat[\label{ea_50}]{%
      \includegraphics[width=0.5\columnwidth]{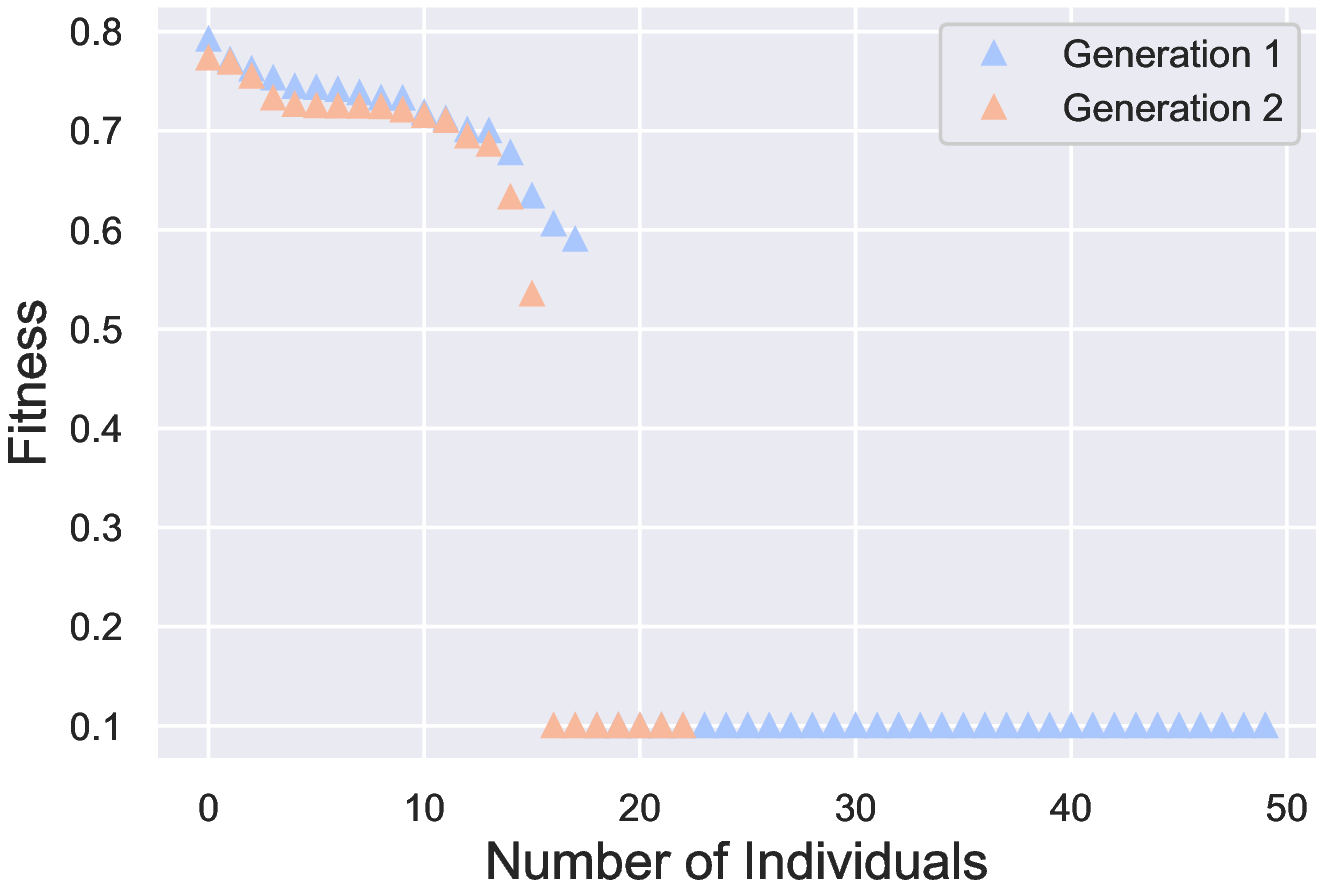}}
  \hfill
  \subfloat[\label{ea_20}]{%
        \includegraphics[width=0.5\columnwidth]{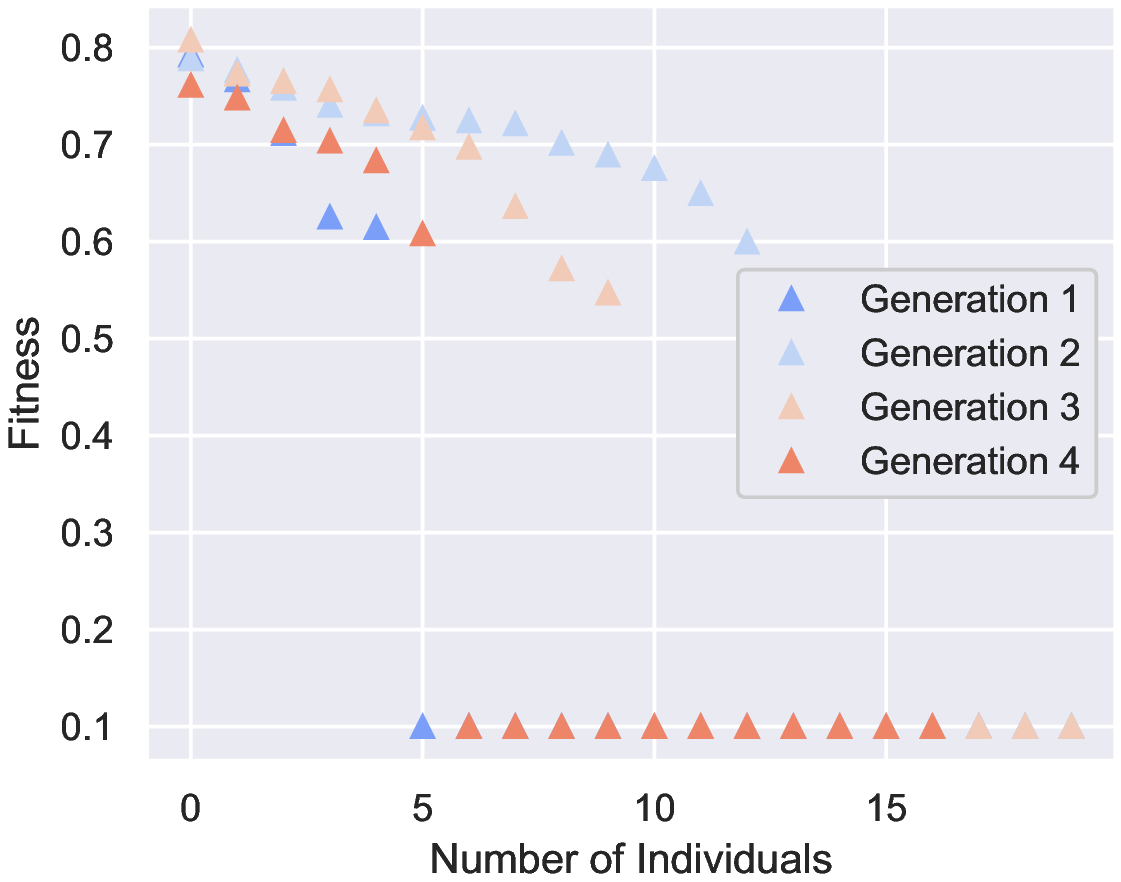}}
  \\
  \subfloat[\label{ea_10}]{%
        \includegraphics[width=0.5\columnwidth]{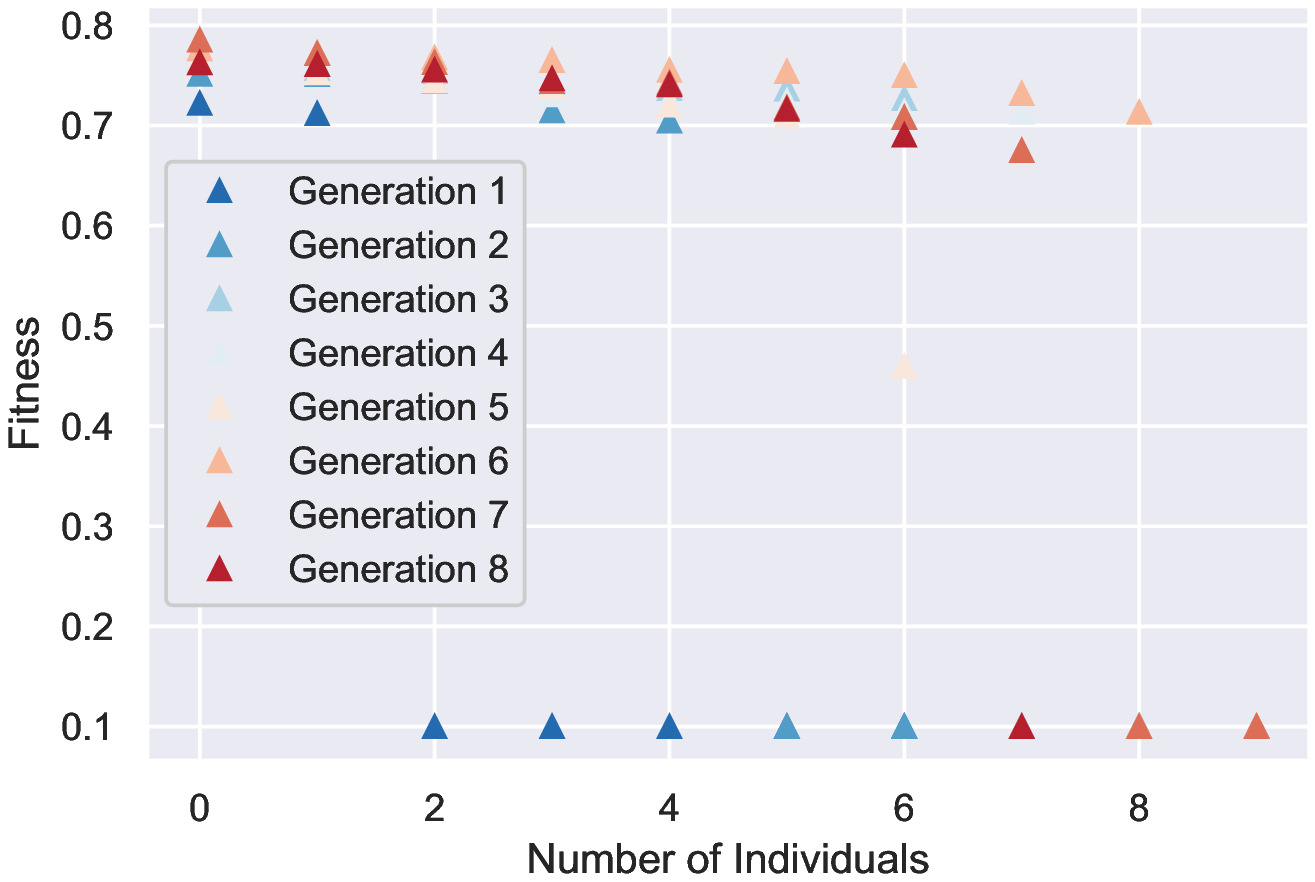}}
\caption{We implemented classical genetic algorithm and ran the experiment with various population sizes and 30 hour time constraint. We sort the fitness of individuals in each generation and plot their fitness. (a)(b)(c) are plots for 50, 20, and 10 population respectively. Since each model is evaluated for 4000 iterations, 30 hour time constraint only allows evaluation of about 75 models. The larger the population size, the less generations will be produced. When the population size is 50, the algorithm stops in the middle of the second generation.}
\label{ea}
\end{figure}

We implemented classical genetic algorithm \cite{Young2015OptimizingAlgorithm} with the same genetic algorithm settings mentioned in the paper, except that instead using a population of size 500, we use much smaller population sizes. Because the fitness score is the model's test accuracy after being trained for 4000 iterations, it takes very long time to evaluate one model. Evaluating the initial 500 models would go far beyond 30 hours. If a 30 hour limit is enforced, the final result would simply be one of the models from the initial population, without any evolution involved. Figure \ref{ea} shows the results produced by the classical genetic algorithm. A 30 hour time constraints only allows evaluation for around 75 models, thus the more individuals in a population, the less generations the population will evolve. If there are 50 individuals in the population, the algorithm will stop in the middle of the second generation, and we do not have much benefits from evolution. 

\begin{figure}[t]
\centering
  \subfloat[\label{lse_30}]{%
      \includegraphics[width=0.5\columnwidth]{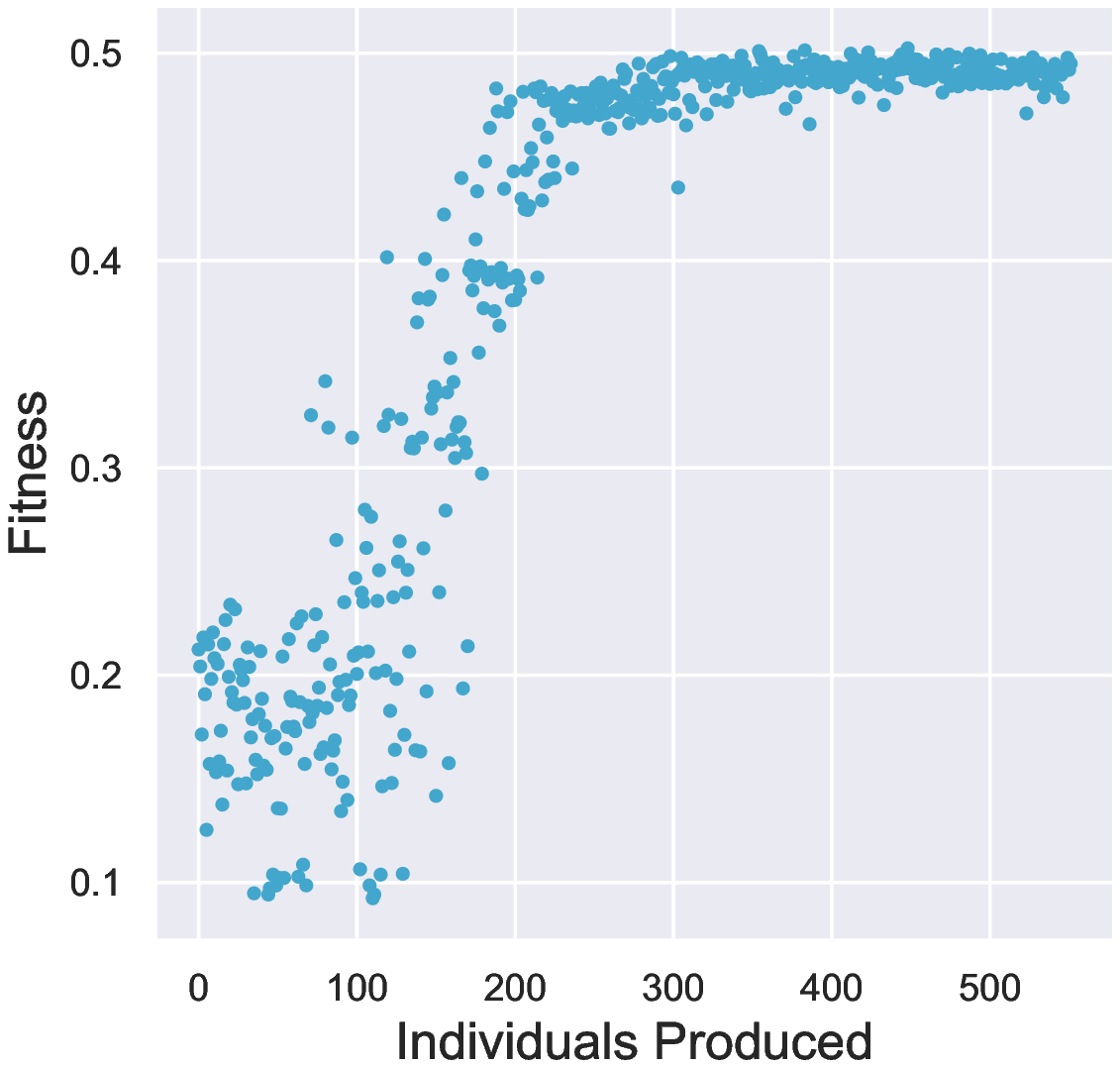}}
  \hfill
  \subfloat[\label{lse_50}]{%
        \includegraphics[width=0.5\columnwidth]{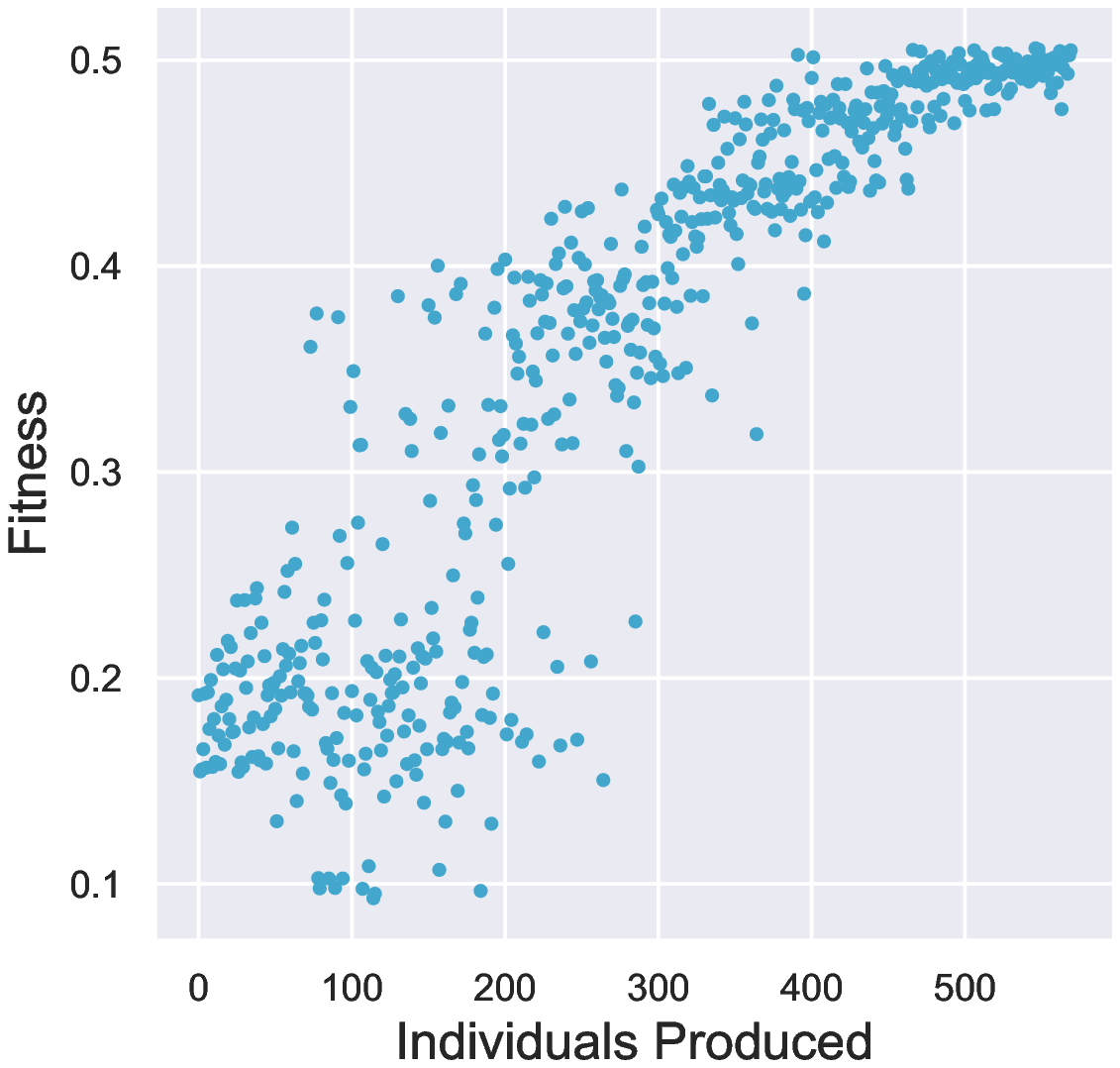}}
  \\
  \subfloat[\label{lse_80}]{%
        \includegraphics[width=0.5\columnwidth]{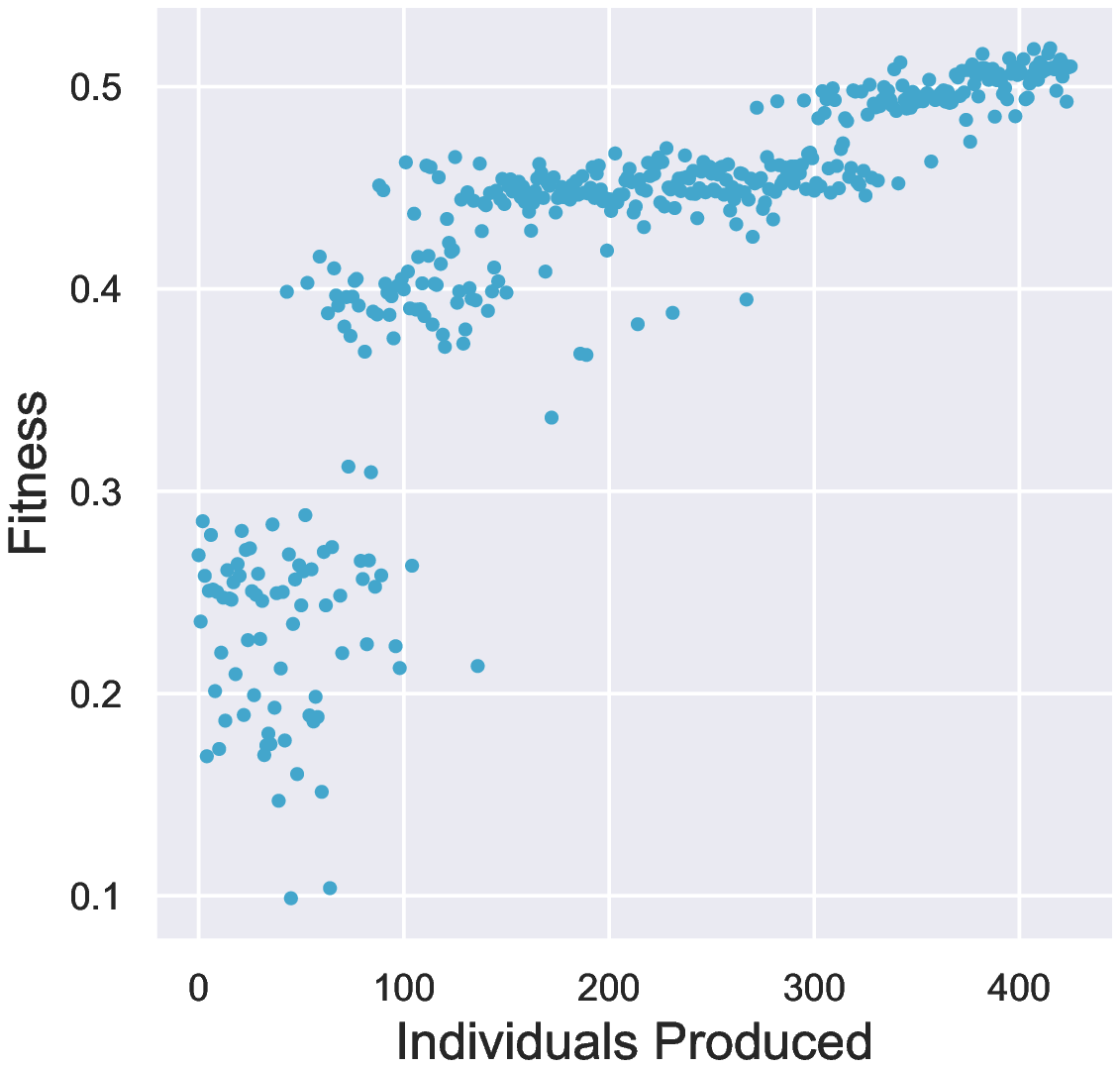}}
  \caption{We implemented large scale evolution on top of a public repository, and ran the experiment with different population setting for 30 hours. (a)(b)(c) are plots for 30, 50, and 80 population respectively. The fitness of the best discovered individuals are slightly above 0.5 when time is restricted to 30 hours on 1 Nvidia Tesla V100 GPU.}
  \label{lse} 
\end{figure}

We implemented large scale evolution \cite{Real2017Large-ScaleClassifiers} on top of a public available implementation \cite{marijnvk2018code}. We ran the implementation with different population size: 30, 50, and 80. The fitness of discovered individuals is shown in Figure \ref{lse}. Large scale evolution can produce good models when given great amount of computing hardware and time. However, when there is limited time and computational hardware available, it is not an ideal algorithm to use. As can be told from Figure \ref{lse_30}, when time is restricted to 1 GPU and 30 hours, the fitness of discovered individuals are relatively low. When the population size is 30, the fitness starts to plateau. This could be due to the lack of variety in the chromosome because the population size is small. When the population size is 80, there is still room for the fitness to improve, but because of a mutation only evolution, it does not grow as fast as our variable length genetic algorithm shown in Figure \ref{evolve_1}.

Table~\ref{tab2} shows the comparison between our method and random search, traditional genetic algorithm \cite{Young2015OptimizingAlgorithm}, and large scale evolution \cite{Real2017Large-ScaleClassifiers}. In our method we sample 20 individuals from the initial search space and evolve for 5 generations in each phase. The best model discovered has 88.92\% test accuracy and the total time is around 25 hours. We ran random search, classical genetic algorithm \cite{Young2015OptimizingAlgorithm}, and large scale evolution \cite{Real2017Large-ScaleClassifiers} with the 30 hour constraint and report the best result found. When we have limited GPU resources, within similar amount of time, large scale evolution gives a model with lowest accuracy. This is because large scale evolution starts with very small models and there are only mutation operations on chromosomes. Without crossover, the evolution process is much slower. Large scale evolution gives very good results when hardware and time is not a problem. However, it is not practical for the average research lab. Classical GA gives a model with relatively high accuracy. This is because the defined search space is much smaller: the number of convolutional layers is fixed to be three according to the paper. It is necessary to fix the layer number because classical GA has fixed length chromosomes. Setting a predefined number of layers is not very practical when we encounter a new problem because we usually do not have much idea of how many layers would be a good fit for the problem. Our method can find better models compared with the other three when there are limited computational resources available.

\begin{table}[htbp]
\begin{center}
\caption{Accuracy Comparison of Discovered models when using similar time (less than 30 hours).}
\begin{tabular}{ccc}
\toprule
  \textbf{Method} &\multicolumn{2}{c}{\textbf{Comparison}} \\
\cline{2-3} 
\textbf{Name} & \textbf{\textit{Accuracy}}& \textbf{\textit{Time(hrs)}}\\

\hline
 Random & 58.66\% & 30\\
 Search & &\\
 \cite{Bergstra2012RandomOptimization} & &\\
 \hline
Classical & 80.75\% & 30\\
GA  & & \\
\cite{Young2015OptimizingAlgorithm} & & \\
\hline
Large Scale & 51.90\% & 30\\
Evolution & &\\
 \cite{Real2017Large-ScaleClassifiers}& &\\
\hline
% ResNet-101 & 9.59\pm0.057 & 95.19\pm0.28 & 8.22\pm0.010 & 95.31\pm0.17\\
%  & &  & (-14.3\%) & (+0.12\%)\\
% \hline
Variable Length & 88.92\% & 24.55\\
 GA (Ours)& &\\
\bottomrule
\end{tabular}
\label{tab2}
\end{center}
\end{table}

\clearpage

\section{Conclusion and Future Work}
\label{sec:conclusion}
In this work we propose to use a variable length genetic algorithm to efficiently find good hyperparameter settings for deep convolutional neural networks. Experiments were performed on CIFAR-10 dataset, and results are compared with random search \cite{Bergstra2012RandomOptimization}, classical genetic algorithm \cite{Young2015OptimizingAlgorithm}, and large scale evolution \cite{Real2017Large-ScaleClassifiers}. Experimental results show that our algorithm can find much better models within a time constraint. 

There are ways of making our algorithm more efficient. For example, right now in each phase all models are trained for 5 epochs for fitness evaluation, even if a model has poor hyperparameter configurations. In the future we can detect these bad configurations in the early stage of fitness evaluation, so we do not spend more computational resources in training them. Other evolutionary algorithms such as ant colony can also be applied to this problem \cite{Wang2015AMinimization}\cite{Wang2008AntNetworks}. We have used fuzzy set theory and evolutionary computing to design a genetic fuzzy routing algorithm in wireless networks \cite{Liu2005AnNetworks}. We have also designed a deep fuzzy neural network for cancer detection \cite{Mudiyanselage2019DeepDetection}. In the future we plan to apply fuzzy set theory to our algorithm to further improve it.

\section*{Acknowledgment}
The authors acknowledge molecular basis of disease (MBD) at Georgia State University for supporting this research, as well as the high performance computing resources at Georgia State University (https://ursa.research.gsu.edu/high-performance-computing) for providing GPU resources. This research is also supported in part by a NVIDIA Academic Hardware Grant.

\bibliographystyle{unsrt}  
% \bibliography{references}  %%% Remove comment to use the external .bib file (using bibtex).


\end{document}